\documentclass[nohyperref]{article}
\pdfoutput=1

\usepackage{amsmath,amsfonts,bm}

\def\eqref#1{equation~\ref{#1}}

\def\1{\bm{1}}

\DeclareMathAlphabet{\mathsfit}{\encodingdefault}{\sfdefault}{m}{sl}
\SetMathAlphabet{\mathsfit}{bold}{\encodingdefault}{\sfdefault}{bx}{n}

\DeclareMathOperator*{\argmax}{arg\,max}

\usepackage{amsmath}
\usepackage{amsfonts}
\usepackage{amssymb}
\usepackage{amsthm}
\usepackage{hyperref}
\usepackage{url}
\usepackage{MnSymbol}
\usepackage{tikz}
\usepackage[all]{xy}
\usepackage[arrowdel]{physics}
\usepackage{algorithm}
\usepackage{algpseudocode}
\usepackage{multirow}
\usepackage{paralist}

\usepackage{microtype}
\usepackage{graphicx}
\usepackage{subfigure}
\usepackage{booktabs} 

\usepackage[accepted]{icml2022}

\usepackage{amsmath}
\usepackage{amssymb}
\usepackage{mathtools}
\usepackage{amsthm}

\usepackage[capitalize,noabbrev]{cleveref}

\theoremstyle{plain}
\newtheorem{theorem}{Theorem}[section]

\theoremstyle{definition}
\newtheorem{definition}[theorem]{Definition}

\theoremstyle{remark}

\newcommand{\m}[1]{$ #1 $}

\usepackage[textsize=tiny]{todonotes}

\icmltitlerunning{Temporal Multiresolution Graph Neural Networks For Epidemic Prediction}

\begin{document}


\twocolumn[
\icmltitle{Temporal Multiresolution Graph Neural Networks For Epidemic Prediction}




\begin{icmlauthorlist}
\icmlauthor{Truong Son Hy}{UChicago}
\icmlauthor{Viet Bach Nguyen}{FPT-AI}
\icmlauthor{Long Tran-Thanh}{Warwick}
\icmlauthor{Risi Kondor}{UChicago}
\end{icmlauthorlist}

\icmlaffiliation{UChicago}{Department of Computer Science, University of Chicago, IL, USA}
\icmlaffiliation{FPT-AI}{FPT Software AI Center, Vietnam}
\icmlaffiliation{Warwick}{Department of Computer Science, University of Warwick, UK}

\icmlcorrespondingauthor{Truong Son Hy}{hytruongson@uchicago.edu}

\icmlkeywords{Graph neural networks, multiresolution learning, timeseries, pandemic prediction}

\vskip 0.3in
]



\printAffiliationsAndNotice{} 

\begin{abstract}
In this paper, we introduce \textit{Temporal Multiresolution Graph Neural Networks} (TMGNN), the first architecture that both learns to construct the multiscale and multiresolution graph structures and incorporates the time-series signals to capture the temporal changes of the dynamic graphs. We have applied our proposed model to the task of predicting future spreading of epidemic and pandemic based on the historical time-series data collected from the actual COVID-19 pandemic and chickenpox epidemic in several European countries, and have obtained competitive results in comparison to other previous state-of-the-art temporal architectures and graph learning algorithms. We have shown that capturing the multiscale and multiresolution structures of graphs is important to extract either local or global information that play a critical role in understanding the dynamic of a global pandemic such as COVID-19 which started from a local city and spread to the whole world. Our work brings a promising research direction in forecasting and mitigating future epidemics and pandemics.
\end{abstract}

\section{Introduction}

Mathematical modeling and simulations of epidemic dynamics play an essential role in understanding and addressing the spreading of infectious diseases in the real world \citep{Kattis2016ModelingEO}. One of the fundamental objects in epidemiology is the acquaintance graph or the social network in which each node of the graph represents an invidual person and each edge represent a human-human interaction \citep{Matt2005}. Because highly contagious diseases including Influenza (i.e. the flu) and COVID-19 can be transmitted airbonne (i.e. respiratory droplets containing the virus) \citep{JAYAWEERA2020109819}, people who are in close contact have a higher risk of spreading the virus from the infected people to the uninfected ones. Thus the acquaintance graph is a reasonable approximation of how the virus spreads in reality \citep{doi:10.1073/pnas.1009094108}. It is important to acquire knowledge from these social networks via graph-based mathematical methods to make prediction on how quick infectious diseases spread over the community \citep{ALGULIYEV2021112}; and to determine the social and public policies accordingly (e.g. to decide which locations should be put into quarantine, to detect which groups of people have a high risk of being infected or need intensive healthcare or should receive the vaccines first, and to examine whether the population achieved some level of herd immunity, etc.). Given a tremendous amount of temporal data collected monthly, weekly, daily and even hourly, there arises a need of large-scale, computationally efficient, and data-driven machine learning models that incorporate both graph information such as human-human interactions and time-series signals such as the status of each person including being uninfected, infected, recovered, vaccinated, etc. In the field of machine learning on graphs, graph neural networks (GNNs) utilizing various ways of generalizing the concept of convolution to graphs have been widely applied to many learning tasks, including modeling physical systems \citep{NIPS2016_3147da8a}, finding molecular representations to estimate quantum chemical computation \citep{Kearnes16}, link prediction on citation graphs \citep{DBLP:conf/iclr/KipfW17}, community detection on social networks \citep{Xing2022}, etc. The most wellknown form of GNNs is message passing neural networks (MPNNs) proposed by \citep{pmlr-v70-gilmer17a}, that are built based on the message passing scheme in which each node propagates and aggregates information, encoded by vectorized messages, to and from its adjacent nodes. MPNNs, based on the mechanism of message exchange in a local neighborhood of nodes, are effective in capturing the local graph structures. But we argue that the fundamental limitation of MPNNs is its lack of a multiresolution and multiscale understanding of the graph that is important to capture both the local and global pictures of an epidemic or a pandemic. One example of multiresolution and multiscale representation on the acquaintance graph is: (i) individual people are represented as single nodes in the graph; (ii) people living in the same city or town form a large cluster of nodes or a supernode, and the amount of movement of people from one city to another is a useful information represented in the edges between supernodes; (iii) finally, a country is a giant supernode containing several cities and towns. Previously, \citep{https://doi.org/10.48550/arxiv.2106.00967} proposed \textit{Multiresolution Graph Networks} (MGN) that constructs multiple resolutions of the input graph via the learning to cluster algorithm in a data-driven manner. Iteratively, MGN uses two modules of graph neural networks on each resolution in which one module learns the graph representation and the another one learns to coarsen the current graph into a number of clusters that further form the graph of the next level of resolution. An important feature of MGN is its flexibility such that the clustering procedure is determined adaptively to the input data rather than being fixed. This flexibility is particularly important for modeling the temporal changes of a global pandemic such as COVID-19 that started with a few cases scattered throughout a local city and finally spread around the globe: the model should pay attention more to the local information in the beginning and gradually pay more attention to the bigger picture when the pandemic progresses. With that motivation, we introduce the use of attention mechanism in order to select the right resolution to make a robust prediction of the current state of a epidemic or pandemic. Putting everything together, we extend the existing MGN architecture to incorporate time-series signals, that results into our newly proposed model \textit{Temporal Multiresolution Graph Neural Networks} (TMGNN). In summary, our contributions are summarized as follows:
\begin{itemize}
\item The use of coarse-grained and hierarchical graph model \citep{https://doi.org/10.48550/arxiv.2106.00967} to capture micro to macro (i.e. local to global) information of a pandemic/epidemic (see Section \ref{sec:mgn}),
\item An attention mechanism to select the right resolution that consequentially predicts the current state or progress of the pandemic/epidemic (see Section \ref{sec:attention}),
\item Temporal architecture incorporating time-series information to make future prediction about the pandemic/epidemic given historical data (see Section \ref{sec:temporal}),
\item Experiments with competitive results on the two temporal datasets of COVID-19 \citep{Panagopoulos_Nikolentzos_Vazirgiannis_2021} and chickenpox \citep{Rozemberczki2021ChickenpoxCI} historical spreading in several countries in Europe (see Section \ref{sec:experiments}).
\end{itemize}
In addition, it is remarkable to note that our multiresolution graph model is computationally efficient, because the size of the graph drops significantly after each level of resolution. The ideal dataset for our proposed model must be the social network of a large number of people including information of acquaintance and human-human interactions. However, due to the lack of such publicly available dataset for the purpose of pandemic modeling, we only applied our method to two existing temporal datasets such as COVID-19 pandemic \citep{Panagopoulos_Nikolentzos_Vazirgiannis_2021} and chickenpox epidemic \citep{Rozemberczki2021ChickenpoxCI} in which each node of the input graph is a geographical region and each edge represents the connection between two neighboring regions. In some sense, these graphs of geographical regions are indeed a low-resolution representation of the social network such that each region is a supernode containing all people living there; and instead of looking at individual human-human interactions, we look at the traffic between regions at a large scale.

\begin{figure*}
\begin{center}
\includegraphics[scale=0.25]{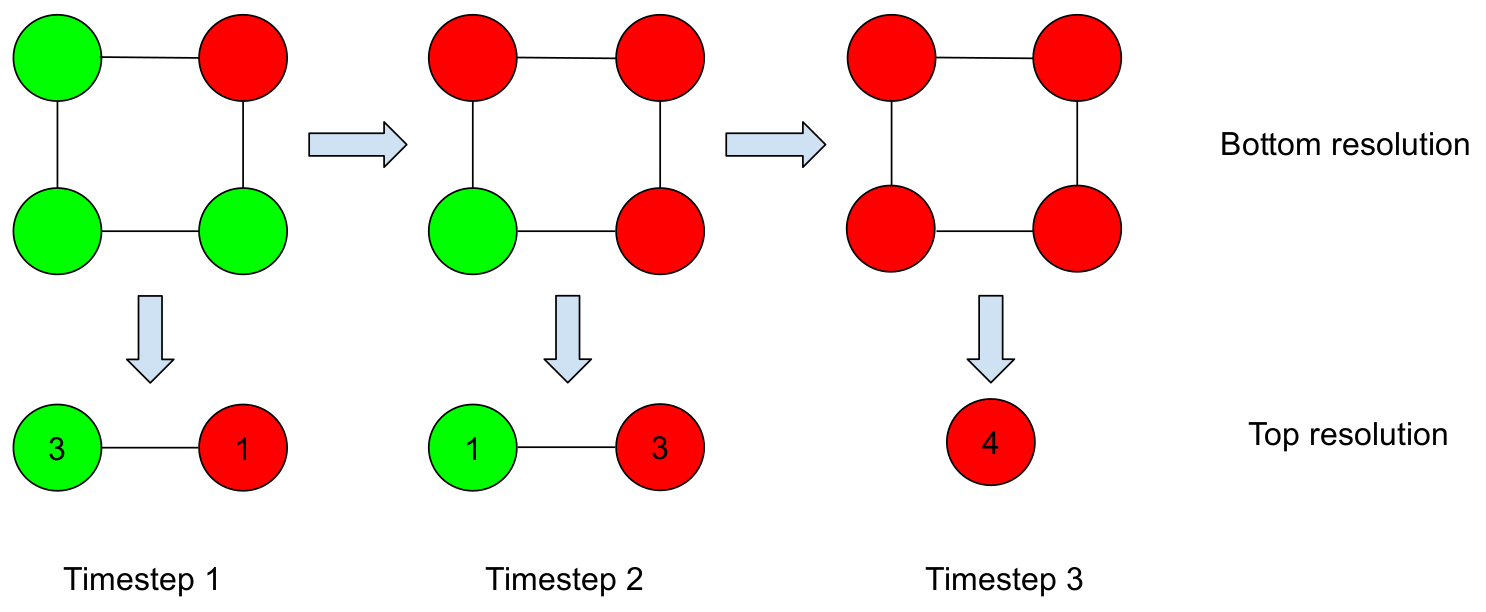}
\end{center}
\caption{\label{fig:clustering} Suppose we are given a interaction network of 4 people as a square graph of 4 nodes and 4 edges. The red color denotes the person gets infected by some virus transmitting via air and the green color denotes otherwise. In this simulation, we consider only 2 levels of resolution: we simply cluster the input graph into 2 clusters (e.g. uninfected and infected). The top row depicts the bottom resolution, while the bottom row depicts the top resolution. At timestep 1, only a single person (at the top right corner) is infected, the input graph is coarsened into 2 supernodes with node labels (3, 1). At timestep 2, because the infected person had interacted with two others, the number of infections increases to 3, and the node labels of the top resolution become (1, 3). At the final timestep, everyone gets infected by the virus, so the coarsened graph contains a single cluster of 4 people while the another cluster is an empty one.}
\end{figure*}
\section{Related work}

In the field of graph analysis, several multiresolution, multiscale and coarse-grained algorithms have been proposed including spectral graph wavelets \citep{HAMMOND2011129}, diffusion wavelets \citep{COIFMAN200653}, graph wavelet transform via multiresolution matrix factorization \citep{hy2021learning}, multilevel graph clustering \citep{4302760} \citep{10.1145/1081870.1081948}, etc. In the recent era of deep learning, graph neural networks (GNNs) as a generalization of the conventional convolution neural networks to graphs have been successful and replacing the traditional graph algorithms \citep{Scarselli09} \citep{Duvenaud2015} \citep{Niepert2016} \citep{pmlr-v70-gilmer17a} \citep{HyEtAl2018} \citep{Kondor2018} \citep{maron2018invariant} \citep{DBLP:journals/corr/abs-2004-03990}. 

Many recent studies have applied deep learning to make predictions about the spreading of COVID-19 pandemic. \citep{CHIMMULA2020109864} proposed the use of time-series based Long Short-Term Memory (LSTM) \citep{Hochreiter97} to predict the number of confirmed COVID-19 cases in Canada. \citep{Kufel_2020} applied ARIMA, a simple autoregressive moving average where the input is the whole time-series of the region up to before the testing day; while \citep{Mahmud_2020} used time-series based PROPHET model\footnote{\url{https://facebook.github.io/prophet/}} as the forecasting model on the same input. \citep{Kapoor2020ExaminingCF} and \citep{10.1093/jamia/ocaa322} introduced the use static and temporal graph neural networks, respectively, on the graph of United States counties to make a prediction for COVID-19 spreading. Others also attempted to blend graph learning models with temporal architectures or recurrent neural networks \citep{Youngjoo2018} \citep{Pareja2020EvolveGCNEG} \citep{10.1145/3341161.3342872} \citep{10.5555/3304222.3304273} \citep{Cornelius2022}; and with reinforcement learning \citep{Meirom2021ControllingGD}. \citep{Rozemberczki2021ChickenpoxCI} established a benchmark for graph neural network architectures in combination with time-series analysis on the dataset of chickenpox cases in Hungary. In addition, \citep{Panagopoulos_Nikolentzos_Vazirgiannis_2021} suggested to use transfer learning in order to translate the knowledge obtained from one country to another.

Our work is unique in the sense that none of the prior works addressed the problem of capturing both local and global information. Obviously, an epidemic or pandemic always starts localized and increases the number of cases over time as more and more people get exposed and infected. Depending on the nature of the virus, the mitigation policies, the weather and many other factors, the speed of spreading can be slow or fast. Each stage of the pandemic needs a different mitigating strategies, thus a forecasting model must take into account both micro and macro signals to make a robust prediction. In this paper, we propose the use of multiresolution graph neural networks to build a hierarchy of resolutions and the attention architecture to select the right resolution at a particular time, that further indicates the corresponding stage of the pandemic.
\section{Method}

\begin{figure*}
\begin{center}
\includegraphics[scale=0.3]{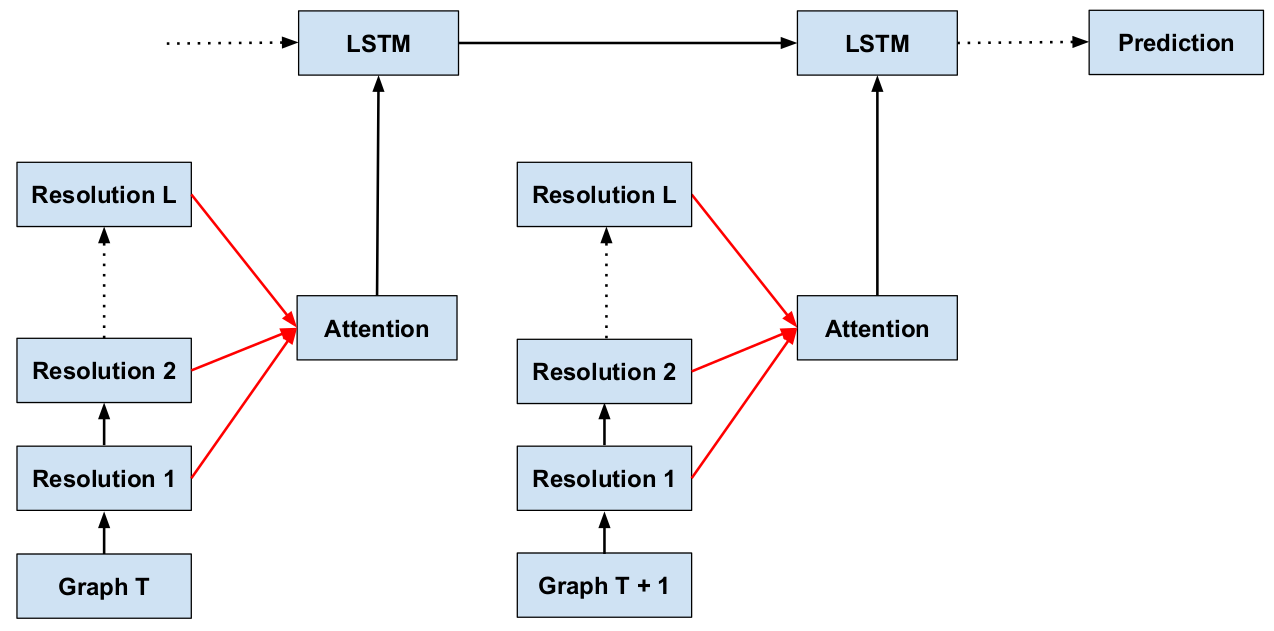}
\end{center}
\caption{\label{fig:temporal-architecture} This diagram, snapshotted at time $T$ and $T + 1$ in the dynamics, depicts the general architecture of Temporal Multiresolution Graph Neural Networks (TMGNN) with the backbone of Long Short-Term Memory (LSTM) \citep{Hochreiter97} or its cousin, Gated Recurrent Unit (GRU) \citep{Cho2014OnTP}. The red arrows denote the attention scores for the multihead attention among resolutions as in section \ref{sec:attention}. The black arrows represent the data flow.}
\end{figure*}

\subsection{Multiresolution Graph Networks} \label{sec:mgn}

\subsubsection{General construction}
Originally, Multiresolution Graph Networks (MGN) and its generative cousin have been proposed by \citep{https://doi.org/10.48550/arxiv.2106.00967}. In this section, we introduce the general construction of MGN. An undirected weighted graph $G = (V, E, A, F_v)$ with node set $V$ and edge set $E$ is represented by an adjacency matrix $A \in \mathbb{R}^{|V| \times |V|}$, where $A_{ij} > 0$ implies an edge between node $v_i$ and $v_j$ with weight $A_{ij}$ (e.g., $A_{ij} \in \{0, 1\}$ in the case of unweighted graph); while node features are represented by a matrix $F_v \in \mathbb{R}^{|V| \times d_v}$.

\begin{definition} \label{def:partition}
A $K$-cluster partition of graph $G$ is a partition of the set of nodes $V$ into $K$ mutually exclusive clusters $V_1, .., V_K$. Each cluster corresponds to an induced subgraph $G_k= G[V_k]$. 
\end{definition}

\begin{definition} \label{def:coarsening}
A coarsening of $G$ is a graph $\tilde{G}$ of $K$ nodes defined by a $K$-cluster partition in which node $\tilde{v}_k$ of $\tilde{G}$ corresponds to the induced subgraph $G_k$. The weighted adjacency matrix $\tilde{A} \in \mathbb{R}^{K \times K}$ of $\tilde{G}$ is 
\[
\tilde{A}_{kk'} = 
\begin{cases} 
\frac{1}{2} \sum_{v_i, v_j \in V_k} A_{ij}, & \mbox{if } k = k', \\ 
\sum_{v_i \in V_k, v_j \in V_{k'}} A_{ij}, & \mbox{if } k \neq k', 
\end{cases}
\]
where the diagonal of $\tilde{A}$ denotes the number of edges inside each cluster, while the off-diagonal denotes the number of edges between two clusters.
\end{definition}

Def.~\ref{def:multiresolution} defines the multiresolution of graph $G$ in a bottom-up manner in which the bottom level is the highest resolution (e.g., $G$ itself) while the top level is the lowest resolution (e.g., $G$ is coarsened into a single node). Def.~\ref{def:mgn} defines a simplified version of MGN.

\begin{definition} \label{def:multiresolution}
An $L$-level coarsening of a graph $G$ is a series of $L$ graphs $G^{(1)}, .., G^{(L)}$ in which 
\begin{enumerate}
\item $G^{(L)}$ is $G$ itself.
\item For $1 \leq \ell \leq L - 1$, $G^{(\ell)}$ is a coarsening graph of $G^{(\ell + 1)}$ as defined in Def.~\ref{def:coarsening}. The number of nodes in $G^{(\ell)}$ is equal to the number of clusters in $G^{(\ell + 1)}$.
\item The top level coarsening $G^{(1)}$ is a graph consisting of a single node, and the corresponding adjacency matrix $A^{(1)} \in \mathbb{R}^{1 \times 1}$. 
\end{enumerate}
\end{definition}

Algorithmically, MGN works in a bottom-up manner as a tree-like hierarchy starting from the highest resolution graph \m{G^{(L)}}, going to the lowest resolution \m{G^{(1)}}. Iteratively, at \m{\ell}-th level, MGN partitions the current graph into \m{K} clusters by running the clustering procedure \m{\bm{c}^{(\ell)}}. Meanwhile, the encoder \m{\bm{e}^{(\ell)}} produces the node latents for this level of resolution. Finally, the pooling network \m{\bm{p}^{(\ell)}} shrinks each cluster into a single node of the next level. In terms of time and space complexity, MGN is more efficient than existing methods in the field because the size of the graph decreases after each resolution.

\begin{definition} \label{def:mgn}
An \m{L}-level Multiresolution Graph Network (MGN)  
of a graph \m{\mathcal{G}} consists of \m{L - 1} 
tuples of three network components 
\m{\{(\bm{c}^{(\ell)}, \bm{e}^{(\ell)}, \bm{p}^{(\ell)})\}_{\ell = 2}^L}. 
The \m{\ell}-th tuple encodes \m{\mathcal{G}^{(\ell)}} and transforms it into a lower resolution graph \m{\mathcal{G}^{(\ell - 1)}} in the higher level. 
Each of these network components has a separate set of learnable parameters \m{(\bm{\theta}_c^{(\ell)}, \bm{\theta}_e^{(\ell)}, \bm{\theta}_p^{(\ell)})}. 
For simplicity, we collectively denote the learnable parameters as \m{\bm{\theta}}, and drop the superscript. 
The network components are defined as follows:
\begin{enumerate}
\item Clustering procedure \m{\bm{c}(G^{(\ell)}; \bm{\theta})}, which partitions graph 
\m{G^{(\ell)}} into \m{K} clusters \m{V_1^{(\ell)}, \ldots, V_K^{(\ell)}}. 
Each cluster is an induced subgraph \m{G_k^{(\ell)}} of 
\m{G^{(\ell)}} with adjacency matrix \m{A_k^{(\ell)}}.

\item Encoder \m{\bm{e}(G^{(\ell)}_k; \bm{\theta})}, which is a 
permutation-equivariant graph neural network that 
takes as input the subgraph \m{G_k^{(\ell)}}, and outputs a set of node latents 
\m{Z^{(\ell)}_k} represented as a matrix of size \m{|V_k^{(\ell)}| \times d_z}.

\item Pooling network \m{\bm{p}(Z^{(\ell)}_k; \bm{\theta})}, which is a permutation-invariant neural network that takes the set of node latents 
\m{Z^{(\ell)}_k} and outputs a single cluster latent \m{\tilde{z}^{(\ell)}_k \in d_z}. 
The coarsening graph \m{G^{(\ell - 1)}} has adjacency matrix \m{A^{(\ell - 1)}} 
built as in Def.~\ref{def:coarsening}, and the corresponding node features 
\m{Z^{(\ell - 1)} = \bigoplus_k \tilde{z}^{(\ell)}_k} represented as a matrix of size \m{K \times d_z}.
\end{enumerate}
\end{definition}

The simplest implementation of the encoder is Message Passing Neural Networks (MPNNs) \citep{pmlr-v70-gilmer17a}. The node embeddings (messages) \m{H_0} are initialized by the input node features: \m{H_0 = F_v}. Iteratively, the messages are propagated from each node to its neighborhood, and then transformed by a combination of linear transformations and non-linearities, e.g.,
$$H_t = \gamma(M_t),$$
$$M_t = D^{-1}AH_{t - 1}W_{t - 1},$$
where \m{\gamma} is a element-wise non-linearity function (e.g., sigmoid, ReLU, tanh, etc.), \m{D_{ii} = \sum_j A_{ij}} is the diagonal matrix of node degrees, and \m{W}s are learnale weight matrices. The output of the encoder is set by messages of the last iteration: \m{Z = H_T}. In this case, $D^{-1}A$ is called the graph Laplacian and we can replace it with the normalized graph Laplacian $I - D^{-1/2} A D^{-1/2}$ in the message passing procedure.

The whole construction of MGN is permutation equivariant with respect to node permutations of \m{G}. In the case of graph property regression, we want MGN to learn to predict a real value $y \in \mathbb{R}$ for each graph $G$ while learning to construct a hierarchical structure of latents and coarsen graphs. The total loss function is
$$
\mathcal{L}_{\text{MGN}}(G, y) = \bigg|\bigg|f\bigg(\bigoplus_{\ell = 1}^L R(Z^{(\ell)})\bigg) - y\bigg|\bigg|_2^2,
$$
where $f$ is a multilayer perceptron, $\oplus$ denotes the vector concatenation, $R$ is a readout function that produces a permutation invariant vector of size $d$ given the latent $Z^{|V^{(\ell)}| \times d}$ at the $\ell$-th resolution.

\subsubsection{Learning to cluster} \label{sec:clustering}

\begin{definition}
A clustering of \m{n} objects into \m{k} clusters is a mapping \m{\pi: \{1, .., n\} \rightarrow \{1, .., k\}} in which \m{\pi(i) = j} if the \m{i}-th object is assigned to the \m{j}-th cluster.
The inverse mapping \m{\pi^{-1}(j) = \{i \in [1, n]: \pi(i) = j\}} gives the set of all objects assigned to the \m{j}-th cluster. 
The clustering is represented by an assignment matrix \m{\Pi \in \{0, 1\}^{n \times k}} such that \m{\Pi_{i, \pi(i)} = 1}.
\end{definition}

The learnable clustering procedure \m{\bm{c}(G^{(\ell)}; \bm{\theta})} is built as follows:
\begin{compactenum}[~~1.] 
\item A graph neural network parameterized by \m{\bm{\theta}} encodes graph \m{G^{(\ell)}} into a first order tensor of \m{K} feature channels \m{\tilde{p}^{(\ell)} \in \mathbb{R}^{|V^{(\ell)}| \times K}}.
\item The clustering assignment is determined by a row-wise maximum pooling operation:
\begin{equation} \label{eq:argmax}
\pi^{(\ell)}(i) = \argmax_{k \in [1, K]} \tilde{p}^{(\ell)}_{i, k}
\end{equation}
that is an equivariant clustering. 
\end{compactenum}
A composition of an equivariant function (e.g., graph net) and an equivariant function (e.g., maximum pooling given in Eq.~\ref{eq:argmax}) is still an equivariant function with respect to the node permutation. Thus, the learnable clustering procedure \m{\bm{c}(G^{(\ell)}; \bm{\theta})} is permutation equivariant. 

In practice, in order to make the clustering procedure differentiable for backpropagation, we replace the maximum pooling in Eq.~\ref{eq:argmax} by sampling from a categorical distribution. Let \m{\pi^{(\ell)}(i)} be a categorical variable with class probabilities \m{p^{(\ell)}_{i, 1}, .., p^{(\ell)}_{i, K}} computed as softmax from \m{\tilde{p}^{(\ell)}_{i, :}}. 
The Gumbel-max trick \citep{Gumbel1954}\citep{NIPS2014_309fee4e}\citep{jang2017categorical} provides a simple and efficient way to draw samples \m{\pi^{(\ell)}(i)}:
\[
\Pi_i^{(\ell)} = \text{one-hot} \bigg( \argmax_{k \in [1, K]} \big[g_{i, k} + \log p^{(\ell)}_{i, k}\big] \bigg),
\]
where \m{g_{i, 1}, .., g_{i, K}} are i.i.d samples drawn from \m{\text{Gumbel}(0, 1)}. 
Given the clustering assignment matrix \m{\Pi^{(\ell)}}, the coarsened adjacency matrix \m{A^{(\ell - 1)}} (see Defs.~\ref{def:partition} and \ref{def:coarsening}) can be constructed as \m{{\Pi^{(\ell)}}^T\! A^{(\ell)} \Pi^{(\ell)}}.

\subsection{Multiresolution Attention} \label{sec:attention}

In this section, we introduce the self-attention mechanism over multiple levels of resolutions. For a given hierarchy of latents $\bm{Z} \triangleq [Z^{(1)}, .., Z^{(L)}]^T \in \mathbb{R}^{L \times d_z}$ where $L$ is the number of resolutions and $Z^{(\ell)}$ is the graph representation at resolution $\ell$-th, produced by MGN (see Def.~\ref{def:mgn}), self-attention transforms $\bm{Z}$ into the output sequence $\bm{X} \triangleq [x_1, .., x_L]^T$ in the following two steps:
\begin{enumerate}
\item The input sequence $\bm{Z}$ is projected into the query matrix $\bm{Q} \triangleq [q_1, .., q_L]^T$, the key matrix $\bm{K} \triangleq [k_1, .., k_L]^T$, and the value matrix $\bm{V} \triangleq [v_1, .., v_L]^T$ via three linear transformations:
$$\bm{Q} = \bm{Z}\bm{W}_Q^T, \ \ \ \ \bm{K} = \bm{Z}\bm{W}_K^T, \ \ \ \ \bm{V} = \bm{Z}\bm{W}_V^T,$$
where $\bm{W}_Q, \bm{W}_K \in \mathbb{R}^{d_k \times d_z}$ and $\bm{W}_V \in \mathbb{R}^{d_v \times d_z}$ are the weight matrices.

\item The output sequence $\bm{X}$ is then computed as follows
\begin{equation}
\bm{X} = \text{Attention}(\bm{Q}, \bm{K}, \bm{V}) = \text{softmax}\bigg( \frac{\bm{Q}\bm{K}^T}{\sqrt{d_k}} \bigg) \bm{V} \triangleq \bm{A}\bm{V},
\label{eq:attention}
\end{equation}
where the softmax function is applied to each row of the matrix $\bm{Q}\bm{K}^T$, and and $\bm{A} \in \mathbb{R}^{L \times L}$ is the attention matrix of $a_{ij}$ attention scores. Equation (\ref{eq:attention}) can be rewritten as
$$x_i = \sum_{j = 1}^L \text{softmax}\bigg( \frac{q_i^T k_j}{\sqrt{d_k}} \bigg) v_j \triangleq \sum_{j = 1}^L a_{ij} v_j.$$
\end{enumerate}
This self-attention is called the scaled dot-product attention or softmax attention. Each output sequence $\bm{X}$ forms an attention head. Let $h$ be the number of heads and $\bm{W}_O \in \mathbb{R}^{hd_v \times hd_v}$ be the projection matrix for the output. In multi-head attention, multiple heads are concatenated to compute the final output defined as follows
$$
\text{MultiHead}(\{\bm{Q}, \bm{K}, \bm{V}\}_{i = 1}^h) = \text{Concat}(\bm{X}_1, .., \bm{X}_h) \bm{W}_O.
$$
Based on the output tensor from the multi-head attention, we contract the dimension of this tensor corresponding to the attention heads, that results into a long vector of $L$ elements in which each corresponds to a resolution. Finally, we apply the Gumbel-softmax trick as in Sec.~\ref{sec:clustering} to convert this vector into an one-hot representation, that allows us to select only one resolution at a particular time. This method is equivalent to detecting the stage of the pandemic over time, whether localized, isolated or uncontrollable. 

\subsection{Temporal Architecture} \label{sec:temporal}

In this section, we put everything together to construct our temporal architecture. Given a sequence of graphs $\big(G_1, G_2, .., G_T\big)$ that correspond to a sequence of timestamps, that can be dates or weeks depending on the dataset. We utilize an Multiresolution Graph Networks (MGN) at each time step $t \in \{1, 2, .., T\}$, to obtain a sequence of representations $\big(Z^{(1)}_t, Z^{(2)}_t, .., Z^{(L)}_t\big)$ for $L$ resolutions of $G_t$. We apply the attention mechanism as in Sec.~\ref{sec:attention} to select one resolution's representation at that time to be fed further into a Long Short-Term Memory (LSTM) backbone \citep{pmlr-v32-graves14}. We expect that the LSTM backbone can capture the long-range temporal dependencies in timeseries and robustly predict the spreading dynamics of the virus based on the encoded graph representation over multiple resolutions. Figure \ref{fig:temporal-architecture} visualizes our temporal architecture.
\section{Experiments} \label{sec:experiments}

Our implementation is done with the temporal library\footnote{\url{https://github.com/benedekrozemberczki/pytorch_geometric_temporal}} as an extension of PyTorch Geometric \citep{Fey:2019wv} \citep{NEURIPS2019_bdbca288}, and is available at {\small\url{https://github.com/bachnguyenTE/temporal-mgn}}. We reuse the original implementation of MGN \citep{https://doi.org/10.48550/arxiv.2106.00967}\footnote{\url{https://github.com/HyTruongSon/MGVAE}}.

\subsection{Chickenpox epidemic in Hungary} \label{sec:Hungary_chickenpox}

\begin{figure*}
\begin{center}
\includegraphics[scale=0.09]{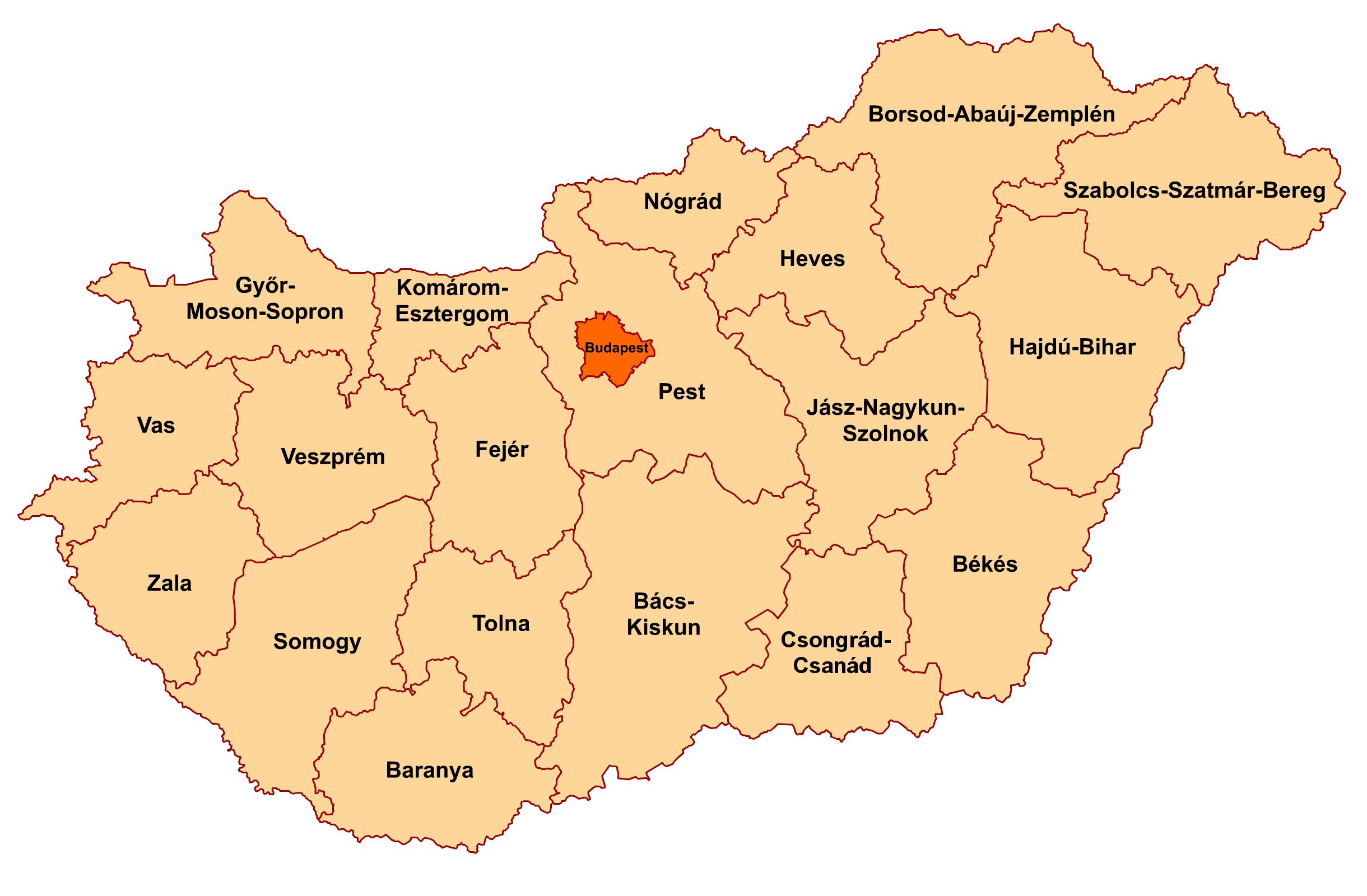}
\includegraphics[scale=0.22]{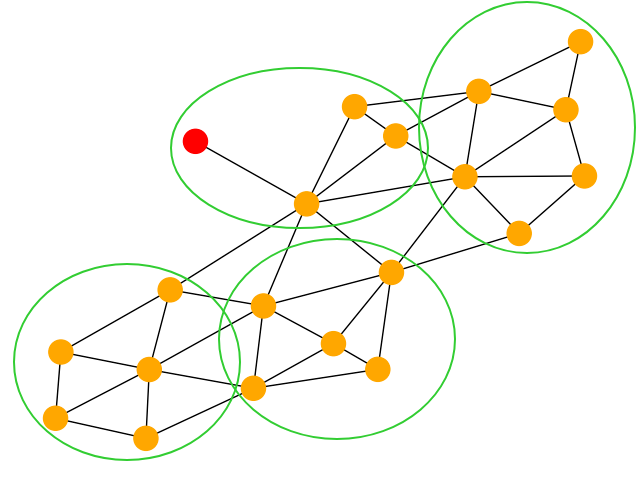}
\includegraphics[scale=0.36]{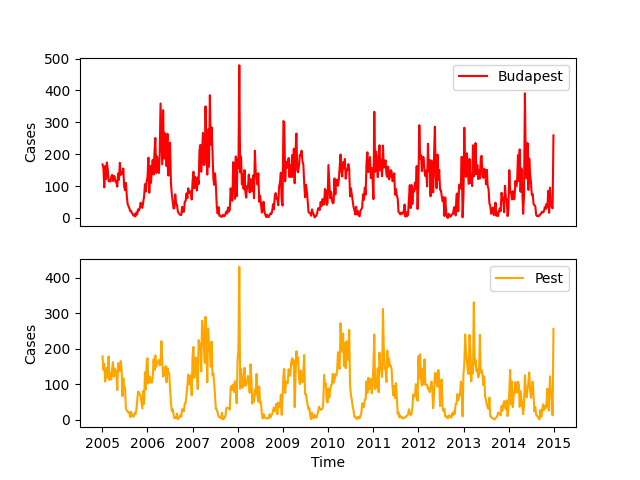}
\end{center}
\caption{\label{fig:Hungary-counties} The left figure is the map of 20 counties in Hungary in 2020, taken from \url{https://en.wikipedia.org/wiki/Counties_of_Hungary}. The middle figure is the corresponding graph in which the red node denotes Budapest, the capital city, and other orange nodes denote the rest of 19 counties. A possible clustering into 4 spatial groups is shown by green circles. The right figure shows the timeseries of chickenpox cases in Budapest and its surrounding county, Pest, are highly correlated. There is clearly a pattern over the period of one year.}
\end{figure*}

Chickenpox is an infection caused by the varicella-zoster virus (VZV). It is highly contagious to those who have not had the disease or been vaccinated against it. Even though vaccines against VZV infection are available \citep{10.1001/jama.287.5.606}, only a small number of countries have national immunization programme \citep{10.1093/bmb/lds019}. In Hungary, parents are routinely recommended to have their children vaccinated but there is no mandatory vaccination against chickenpox. Hungarian physicians have to report each case to the local centre of epidemiology which are then aggregated and publicly presented weekly for each of the 20 counties of the country \citep{Karsai2020}. \textit{Chickenpox Cases in Hungary} is a novel spatiotemporal dataset which can be uesd to benchmark the forecasting performance of graph neural network architectures \citep{Rozemberczki2021ChickenpoxCI}. The dataset consists of:
\begin{itemize}
\item A spatial graph that describes the spatial connectivity of the 20 counties (i.e. nodes) including 19 counties and the capital Budapest with 61 edges (i.e. connections) between them (see Fig.~\ref{fig:Hungary-counties}).
\item The country level timeseries, collected from the digital version of the \textit{Hungarian Epidemiological Info}\footnote{\url{http://www.oek.hu/}} bulletin, that describe the weekly number of chickenpox cases reported by Hungarian general practitioners from January of 2005 to January of 2015, with totally more than 500 entries for each county without any missingness (see Fig.~\ref{fig:Hungary-counties}).
\end{itemize}

In our experiment of TMGNN for this dataset, we use the following setup. We have three levels of resolution with the number of nodes/clusters per resolution as 20, 8 and 1. In each resolution, we execute 4 layers of message passing with the size of the message per node is 24. Our model uses 8 temporal lags as the input node features. For the attention mechanism to select one resolution at a given time, the number of heads in the multihead attention ranges from 1 to 4. The training \& testing dataset ratio is 80\%/20\%. We train our model for 150 epochs by Adam optimizer \citep{Kingma2015} with the initial learning rate as $0.001$. All other setting details are similar to \citep{Rozemberczki2021ChickenpoxCI}. 

In table \ref{table:chickenpox}, we report the average mean squared error with standard deviation for forecasting horizon of 40 weeks calculated from 10 experimental runs, in comparison with several other temporal graph neural networks and recurrent neural networks. Our proposed TMGNN model outperforms the best result reported in \citep{Rozemberczki2021ChickenpoxCI} by 11\%.

\begin{table}
\begin{center}
\begin{tabular}{|l|c|}
\hline
\multicolumn{2}{|c|}{Hungary} \\
\hline
GConvLSTM \citep{Youngjoo2018} & 1.221 $\pm$ 0.010 \\
GConvGRU \citep{Youngjoo2018} & 1.117 $\pm$ 0.002 \\
Evolve GCN-O \citep{Pareja2020EvolveGCNEG} & 1.120 $\pm$ 0.003 \\
Evolve GCN-H \citep{Pareja2020EvolveGCNEG} & 1.115 $\pm$ 0.013 \\
DynGRAE \citep{10.1145/3308560.3316581} & 1.112 $\pm$ 0.010 \\
STGCN \citep{10.5555/3304222.3304273} & 1.118 $\pm$ 0.005 \\
DCRNN \citep{li2018diffusion} & 1.119 $\pm$ 0.002 \\
\hline
TMGNN (ours) & \textbf{0.990 $\pm$ 0.003} \\
\hline
\end{tabular}
\end{center}
\caption{\label{table:chickenpox} The average test mean squared error with standard deviations obtained over 40 weeks long forecasting horizons in Hungary calculated from a 10 experimental runs with different random seeds. The baseline results are taken from \citep{Rozemberczki2021ChickenpoxCI}. The best result is marked boldly.}
\end{table}
\subsection{COVID-19 pandemic in Europe}


COVID-19 is disease caused by severe acute respiratory syndrome coronavirus 2 or SARS-Cov-2, a newly discovered virus that is closely related to bat coronaviruses \citep{doi:10.1056/NEJMe2001126}, pangolin coronaviruses \citep{ZHANG20201346} and SARS-CoV \citep{SUN2020483}. The novel virus was first detected from an outbreak in Wuhan, China in December 2019. The World Health Organization (WHO) declared a Public Health Emergency of International Concern on January 30, 2020 and a pandemic on March 11, 2020. At the time of this article was written, the COVID-19 pandemic was still an ongoing global pandemic. Since the early outbreak, global cooperation among doctors, medical staffs, scientists and engineers has been facilitated through an unprecedented amount of information and data. In the scope of Data for Good program \footnote{\url{https://dataforgood.facebook.com/dfg/tools}}, Facebook Inc.~has released many datasets to help researchers better understand the dynamics of COVID-19 and predict the spread of this disease \citep{Maas2019}. In this experiment, the dataset contains measures of human mobility between regions of the Nomenclature of Territorial Units for Statistics 3 (NUTS3), a geocode standard for referencing the subdivisions of member states of the European Union (EU). The raw timeseries data, collected and aggregated from mobile phones that have Facebook application installed and Location History setting enabled, includes three recordings per day (i.e., midnight, morning and afternoon) indicating the number of people travelling from on region to another during that period of time in the day. We reuse the preprocessed timeseries data from \citep{Panagopoulos_Nikolentzos_Vazirgiannis_2021}\footnote{\url{https://github.com/geopanag/pandemic_tgnn}} in which three values per day are further aggregated into a single number for representing the mobility between two regions. It has been known that the probability that people living in a region are infected by the virus increases given more people moving in and out from that region \citep{doi:10.1073/pnas.0510525103} \citep{Soriano_Pa_os_2020}, that suggests the use of message passing scheme and graph neural networks can be effective in capturing the amount of mobility and predicting the number of infections in each region with high accuracies. However, previous works exploiting graph structures of regions lacked the ability to obtain a bigger picture over multiple regions. Our proposal of using multiscale and multiresolution graph networks along with an attention mechanism to robustly choosing the right resolution addresses this limitation. Our model can capture the local signals and aggregate into global signals, and is adaptive to predict the new stage of the ongoing pandemic. We support our claim by numerical results on the historical data of several member countries of the EU including England, France, Italy and Spain in tables \ref{table:England-Covid-19}, \ref{table:France-Covid-19}, \ref{table:Italy-Covid-19} and \ref{table:Spain-Covid-19}, respectively. We compare our model TMGNN with the same baselines as in \citep{Panagopoulos_Nikolentzos_Vazirgiannis_2021}:
\begin{itemize}
\vspace{-0.25cm}
\item Statistical analysis including: (i) AVG, the average number of cases for the specific region; (ii) AVG\_WINDOW, the average number of cases in the past $d$ days for the specific region (i.e. moving average); and (iii) LAST\_DAY, the number of cases in the previous day is used as the prediction for the next days.
\item Timeseries models without graph topology: (i) LSTM \citep{CHIMMULA2020109864}, a simple Long Short-Term Memory architecture taking the input as the sequence of new cases in a region for the previous week; (ii) ARIMA \citep{Kufel_2020}, a simple autoregressive moving average model taking the input as the whole sequence in the region; and (iii) PROPHET \citep{Mahmud_2020}, similar to ARIMA but with more types/features of the timeseries.
\item TL\_BASE, MPNN and MPNN+LSTM \citep{Panagopoulos_Nikolentzos_Vazirgiannis_2021}: Baseline models using the conventional message passing neural networks with LSTM for processing timeseries.
\item MGN: Furthermore, we apply the multiresolution graph networks \citep{https://doi.org/10.48550/arxiv.2106.00967}.
\vspace{-0.25cm}
\end{itemize}
Our training and hyperparameter settings are similar to the Hungarian chickenpox experiment in Sec.~\ref{sec:Hungary_chickenpox}. We use 64 dimensional node messages, and four levels of resolution including the bottom resolution as the input graph and then a hierarchy of three coarsen graphs of sizes 32, 16 and 8, respectively. We evaluate the performance of a model by comparing the prediction of the total number of cases in each region verus ground truth from the historical data, in a test set with the mean average error metric from \citep{Panagopoulos_Nikolentzos_Vazirgiannis_2021} defined as follows:
$$\text{MAE}(\bm{\hat{y}}, \bm{y}) = \frac{1}{n \times d} \sum_{t = T + 1}^{T + d} \sum_{v \in V} |\hat{y}_v^{(t)} - y_v^{(t)}|,$$
where $G = (V, E)$ denotes the graph of a country in which $V$ denotes the set of regions (i.e. nodes) and $E$ denotes the set of inter-regional connections (i.e. edges), $\hat{y}_v^{(t)}$ is the prediction of the model at time $t$ for the $v$-th region and $y_v^{(t)}$ is the corresponding ground truth, and finally $d$ is the number of days that we need to predict ahead. In our experiments, $d$ are set to be 3 days, 7 days and 14 days and the corresponding results are shown in three columns of each table. We outperform all state-of-the-art methods in 6 out of 12 experiments. 

\section{Conclusion}

For the purpose of predicting the dynamics of a pandemic or an epidemic, we propose a novel temporal architecture of graph neural networks that combines both graph information of human interactions and timeseries signals of human mobility. Our novelty comes from a mechanism to build the multiresoluiton or multiscale representation of the graph in a data-driven manner and using attention to select the right resolution at a given time. This combination of ideas allows our proposed model to be adaptive and robust in estimating the current and future stage of the COVID-19 pandemic. We support our claim by strong numerical results based on real historical data. Our temporal multiresolution graph model is promising to be applied into a larger-scale data and tackling more challenging predictive tasks.

\begin{table}
\vspace{0.36cm}
\begin{center}
\scalebox{1}{
\begin{tabular}{|l|c|c|c|}
\hline
\multicolumn{4}{|c|}{England} \\
\hline
\textbf{Model} & \textbf{3 days} & \textbf{7 days} & \textbf{14 days} \\
\hline
AVG & 9.75 & 9.99 & 10.09 \\
LAST DAY & 7.11 & 7.62 & 8.66 \\
AVG\_WINDOW & 6.52 & 7.34 & 8.54 \\
LSTM & 9.11 & 8.97 & 9.10 \\
ARIMA & 13.77 & 14.55 & 15.65 \\
PROPHET & 10.58 & 12.25 & 16.24 \\
TL\_BASE & 9.65 & 12.30 & 13.48 \\
MPNN & 6.36 & 6.86 & 8.13 \\
MPNN+LSTM & 6.41 & 6.67 & 7.02 \\
MGN & 6.68 & 7.37 & 8.74 \\
\hline
TMGNN (ours) & \textbf{6.26} & \textbf{6.55} & \textbf{6.80} \\
\hline
\end{tabular}
}
\end{center}
\caption{\label{table:England-Covid-19} The average error of predictions for up to next 3, 7 and 14 days in England.}
\end{table}
\begin{table}
\begin{center}
\scalebox{1}{
\begin{tabular}{|l|c|c|c|}
\hline
\multicolumn{4}{|c|}{France} \\
\hline
\textbf{Model} & \textbf{3 days} & \textbf{7 days} & \textbf{14 days} \\
\hline
AVG & 8.50 & 8.55 & 8.55 \\
LAST DAY & 7.47 & 7.37 & 8.03 \\
AVG\_WINDOW & \textbf{6.04} & 6.40 & 7.24 \\
LSTM & 8.08 & 8.13 & 7.91 \\
ARIMA & 10.72 & 10.53 & 10.91 \\
PROPHET & 10.34 & 11.56 & 14.61 \\
TL\_BASE & 7.67 & 9.21 & 12.27 \\
MPNN & 6.16 & \textbf{5.99} & \textbf{6.93} \\
MPNN+LSTM & 6.39 & 7.21 & 7.36 \\
MGN & 6.65 & 6.61 & 7.66 \\
\hline
TMGNN (ours) & 6.39 & 7.35 & 7.51 \\
\hline
\end{tabular}
}
\end{center}
\caption{\label{table:France-Covid-19} The average error of predictions for up to next 3, 7 and 14 days in France.}
\end{table}
\begin{table}
\begin{center}
\begin{tabular}{|l|c|c|c|}
\hline
\multicolumn{4}{|c|}{Italy} \\
\hline
\textbf{Model} & \textbf{3 days} & \textbf{7 days} & \textbf{14 days} \\
\hline
AVG & 21.38 & 22.23 & 23.09 \\
LAST DAY & 17.40 & 18.49 & 20.69 \\
AVG\_WINDOW & 15.17 & 16.81 & 19.45 \\
LSTM & 22.94 & 23.17 & 23.12 \\
ARIMA & 35.28 & 37.23 & 39.65 \\
PROPHET & 24.86 & 27.39 & 33.07 \\
TL\_BASE & 19.12 & 23.44 & 24.89 \\
MPNN & \textbf{14.39} & \textbf{15.47} & 17.88 \\
MPNN+LSTM & 15.56 & 16.41 & 17.25 \\
MGN & 15.33 & 16.73 & 19.22 \\
\hline
TMGNN (ours) & 15.09 & 15.62 & \textbf{16.38} \\
\hline
\end{tabular}
\end{center}
\caption{\label{table:Italy-Covid-19} The average error of predictions for up to next 3, 7 and 14 days in Italy.}
\end{table}
\begin{table}
\begin{center}
\begin{tabular}{|l|c|c|c|}
\hline
\multicolumn{4}{|c|}{Spain} \\
\hline
\textbf{Model} & \textbf{3 days} & \textbf{7 days} & \textbf{14 days} \\
\hline
AVG & 45.10 & 45.87 & 47.63 \\
LAST DAY & 33.58 & 37.06 & 43.63 \\
AVG\_WINDOW & \textbf{32.19} & 36.06 & 42.79 \\
LSTM & 51.44 & 49.89 & 47.26 \\
ARIMA & 40.49 & 41.64 & 46.22 \\
PROPHET & 54.76 & 62.16 & 79.42 \\
TL\_BASE & 42.25 & 52.29 & 59.68 \\
MPNN & 35.83 & 38.51 & 44.25 \\
MPNN+LSTM & 33.35 & 34.47 & 35.31 \\
MGN & 38.85 & 43.65 & 52.23 \\
\hline
TMGNN (ours) & 33.13 & \textbf{34.12} & \textbf{35.04} \\
\hline
\end{tabular}
\end{center}
\caption{\label{table:Spain-Covid-19} The average error of predictions for up to next 3, 7 and 14 days in Spain.}
\end{table}

\bibliography{temporal_mgn}
\bibliographystyle{icml2022}




\end{document}